\documentclass[11pt,a4paper]{article}

\usepackage[hyperref]{acl2018}

\aclfinalcopy

\usepackage{ifxetex}
\ifxetex
  \usepackage{fontspec}
\else
  \usepackage[T1]{fontenc}
  \usepackage[utf8]{inputenc}
\fi

\usepackage{times}
\usepackage{latexsym}

\usepackage{xspace}
\newcommand*{\eg}{e.g.\@\xspace}
\newcommand*{\ie}{i.e.\@\xspace}

\makeatletter
\newcommand*{\etc}{%
	\@ifnextchar{.}%
	{etc}%
	{etc.\@\xspace}%
}
\makeatother

\usepackage{url}
\usepackage{graphicx}
\usepackage{booktabs}
\usepackage[capitalise,noabbrev]{cleveref}
\usepackage{tabularx}

\usepackage{url}
\usepackage{paralist}
\usepackage{comment}
\usepackage{multirow}

\usepackage{listings}
\lstdefinestyle{nonumbers}{numbers=none}

\usepackage{xcolor}

\colorlet{punct}{red!60!black}
\definecolor{delim}{RGB}{20,105,176}
\colorlet{numb}{magenta!60!black}

\usepackage{tikz}
\usetikzlibrary{calc,trees,positioning,arrows,chains,shapes.geometric,%
  decorations.pathreplacing,decorations.pathmorphing,shapes,%
  matrix,shapes.symbols,fit,decorations,arrows.meta,%
  fit,shapes.misc}
\definecolor{nice-red}{HTML}{E41A1C}
\colorlet{dark-red}{nice-red!80!black}
\definecolor{nice-orange}{HTML}{FF7F00}
\colorlet{dark-orange}{orange!85!black}
\definecolor{nice-yellow}{HTML}{FFC020}
\definecolor{nice-green}{HTML}{4DAF4A}
\definecolor{nice-blue}{HTML}{377EB8}
\definecolor{nice-purple}{HTML}{984EA3}

\DeclareFixedFont{\ttb}{T1}{txtt}{bx}{n}{12} 
\DeclareFixedFont{\ttm}{T1}{txtt}{m}{n}{12}  

\usepackage{color}
\definecolor{deepblue}{rgb}{0,0,0.5}
\definecolor{deepred}{rgb}{0.6,0,0}
\definecolor{deepgreen}{rgb}{0,0.5,0}

\newcommand*{\affaddr}[1]{#1} 
\newcommand*{\affmark}[1][*]{\textsuperscript{#1}}

\newcommand{\jtr}{{\textsc{Jack}}}

\aclfinalcopy

\title{Jack the Reader -- A Machine Reading Framework}

\author{%
\bf Dirk Weissenborn\affmark[1],
Pasquale Minervini\affmark[2],
Tim Dettmers\affmark[3],
Isabelle Augenstein\affmark[4],\\
\bf Johannes Welbl\affmark[2],
Tim Rockt{\"a}schel\affmark[5],
Matko Bo\v{s}njak\affmark[2],
Jeff Mitchell\affmark[2],\\
\bf Thomas Demeester\affmark[6],
Pontus Stenetorp\affmark[2],
Sebastian Riedel\affmark[2]\\
\affaddr{\affmark[1]German Research Center for Artificial Intelligence (DFKI), Germany}\\
\affaddr{\affmark[2]University College London, United Kingdom}\\
\affaddr{\affmark[3]Università della Svizzera italiana, Switzerland}\\
\affaddr{\affmark[4]University of Copenhagen, Denmark}\\
\affaddr{\affmark[5]University of Oxford, United Kingdom}\\
\affaddr{\affmark[6]Ghent University - imec, Ghent, Belgium}
}

\date{}

\begin{document}

\maketitle

\begin{abstract}
Many Machine Reading and Natural Language Understanding tasks require reading supporting text in order to answer questions.
For example, in Question Answering, the supporting text can be newswire or Wikipedia articles; in Natural Language Inference, premises can be seen as the supporting text and hypotheses as questions.
Providing a set of useful primitives operating in a single framework of related tasks would allow for expressive modelling, and easier model comparison and replication.
To that end, we present Jack the Reader ({\jtr}), a framework for Machine Reading that allows for quick model prototyping by component reuse, evaluation of new models on existing datasets as well as integrating new datasets and applying them on a growing set of implemented baseline models. 
{\jtr} is currently supporting (but not limited to) three tasks: Question Answering, Natural Language Inference, and Link Prediction.
It is developed with the aim of increasing research efficiency and code reuse.
\end{abstract}

\section{Introduction} \label{sec:introduction}

Automated reading and understanding of textual and symbolic input, to a degree that enables question answering, is at the core of Machine Reading (\emph{MR}).
A core insight facilitating the development of MR models is that most of these tasks can be cast as an instance of the Question Answering (\emph{QA}) task: an input can be cast in terms of \emph{question}, \emph{support documents} and \emph{answer candidates}, and an output in terms of \emph{answers}.
For instance, in case of Natural Language Inference (\emph{NLI}), we can view the hypothesis as a multiple choice question about the underlying premise (support) with predefined set of specific answer candidates (entailment, contradiction, neutral).
Link Prediction (\emph{LP}) -- a task which requires predicting the truth value about facts represented as \textit{(subject, predicate, object)}-triples -- can be conceived of as an instance of QA (see \cref{sec:supported_tasks} for more details).
By unifying these tasks into a single framework, we can facilitate the design and construction of multi-component MR pipelines.
There are many successful frameworks such as
\textsc{Stanford CoreNLP}~\citep{manning-EtAl:2014:P14-5}, \textsc{NLTK}~\citep{bird2009natural}, and \textsc{spaCy}\footnote{\url{https://spacy.io}} for NLP, \textsc{Lucene}\footnote{\url{https://lucene.apache.org}} and \textsc{Solr}\footnote{\url{http://lucene.apache.org/solr/}} for Information Retrieval, and \textsc{scikit-learn}\footnote{\url{http://scikit-learn.org}}, \textsc{PyTorch}\footnote{\url{http://pytorch.org/}} and \textsc{TensorFlow}~\citep{tensorflow2015-whitepaper} for general Machine Learning (\textit{ML}) with a special focus on Deep Learning (\emph{DL}), among others.
All of these frameworks touch upon several aspects of Machine Reading, but none of them offers dedicated support for modern MR pipelines. Pre-processing and transforming MR datasets into a format that is usable by a MR model as well as implementing common architecture building blocks all require substantial effort which is not specifically handled by any of the aforementioned solutions.
This is due to the fact that they serve a different, typically much broader purpose.
In this paper, we introduce Jack the Reader ({\jtr}), a reusable framework for MR.
It allows for the easy integration of novel tasks and datasets by exposing a set of high-level primitives and a common data format.
For supported tasks it is straight-forward to develop new models without worrying about the cumbersome implementation of training, evaluation, pre- and post-processing routines.
Declarative model definitions make the development of QA and NLI models using common building blocks effortless.
{\jtr} covers a large variety of datasets, implementations and pre-trained models on three distinct MR tasks and supports two ML backends, namely \textsc{PyTorch} and \textsc{TensorFlow}. 
Furthermore, it is easy to train, deploy, and interact with MR models, which we refer to as \emph{readers}.
\section{Related Work} \label{sec:related}
Machine Reading requires a tight integration of Natural Language Processing and Machine Learning models.
General NLP frameworks include \textsc{CoreNLP}~\cite{manning-EtAl:2014:P14-5}, \textsc{NLTK}~\cite{bird2009natural}, \textsc{OpenNLP}\footnote{\url{https://opennlp.apache.org}} and \textsc{spaCy}. %
All these frameworks offer pre-built models for standard NLP preprocessing tasks, such as tokenisation, sentence splitting, named entity recognition and parsing.
\textsc{GATE}~\cite{cunningham-EtAl:2002:ACL} and \textsc{UIMA}~\cite{ferrucci2004uima} are toolkits that allow quick assembly of baseline NLP pipelines, and visualisation and annotation via a Graphical User Interface.
\textsc{GATE} can utilise \textsc{NLTK} and \textsc{CoreNLP} models and additionally enable development of rule-based methods using a dedicated pattern language.
\textsc{UIMA} offers a text analysis pipeline which, unlike \textsc{GATE}, also includes retrieving information, but does not offer its own rule-based language.
It is further worth mentioning the Information Retrieval frameworks \textsc{Apache Lucene} and \textsc{Apache Solr} which can be used for building simple, keyword-based question answering systems, but offer no ML support.
Multiple general machine learning frameworks, such as \textsc{scikit-learn}~\cite{pedregosa2011scikit}, \textsc{PyTorch}, \textsc{Theano}~\cite{2016arXiv160502688short} and \textsc{TensorFlow}~\cite{tensorflow2015-whitepaper}, among others, enable quick prototyping and deployment of ML models.
However, unlike {\jtr}, they do not offer a simple framework for defining and evaluating MR models.
The framework closest in objectives to {\jtr} is \textsc{AllenNLP}~\cite{gardner2017allennlp}, which is a research-focused open-source NLP library built on \textsc{PyTorch}.
It provides the basic low-level components common to many systems in addition to pre-assembled models for standard NLP tasks, such as coreference resolution, constituency parsing, named entity recognition, question answering and textual entailment.
In comparison with \textsc{AllenNLP}, {\jtr} supports both \textsc{TensorFlow} and \textsc{PyTorch}.
Furthermore, {\jtr} can also learn from Knowledge Graphs (discussed in \cref{sec:supported_tasks}), while \textsc{AllenNLP} focuses on textual inputs.
Finally, {\jtr} is structured following a modular architecture, composed by input-, model-, and output modules, facilitating code reuse and the inclusion and prototyping of new methods.
\section{Overview} \label{sec:overview}
\begin{figure}[t]
	\centering
	\resizebox{\columnwidth}{!}{
	\begin{tikzpicture}
	    \node[ultra thick, draw, rectangle, rounded corners, fill=nice-blue!30, text width=2.1cm, minimum height=2.2cm] (mr) {{\large \sc JTReader}\\
	    {\small • Save, Load}\\
	    {\small • Setup, Train}};
	    
	    \node[thick, draw, left = of mr, rectangle split, rectangle split parts=3, rounded corners, text width=3.4cm] (modules) {
	    
	    {\sc Input}\\[-0.1cm]
	    {\small • Vocabulary building}\\[-0.1cm]
	    {\small • Embeddings}\\[-0.1cm]
	    {\small • Batching}\\[-0.1cm]
	    {\small • Data to tensors}
	    
	    \nodepart{two}
	    {\sc Model}\\
	    {\small • TensorFlow/PyTorch}
	    
	    \nodepart{three}
	    {\sc Output}
	    \\{\small • Human-readable output}};
	    
	    \draw[thick, dashed] ([yshift=-0.1cm]modules.north east) -- (mr.north west);
	    \draw[thick, dashed] ([yshift=0.1cm]modules.south east) -- (mr.south west);
	    
	    \node[right = 2cm of mr] (dummy) {};
	    \node[very thick, rectangle, draw, rounded corners, above = 0.25cm of dummy, fill=nice-orange!50, dotted, minimum width=2.2cm, minimum height=0.75cm] (support) {Support};
	    \node[thick, rectangle, draw, rounded corners, above = 0.25cm of support, fill=nice-red!50, minimum width=2.2cm, minimum height=0.75cm] (question) {Question};
	    
	    \node[very thick, rectangle, draw, rounded corners, below = 0.25cm of dummy, fill=nice-yellow!50, dotted, minimum width=2.2cm, minimum height=0.75cm] (candidate) {Candidates};
	    \node[thick, rectangle, draw, rounded corners, below = 0.25cm of candidate, fill=nice-green!50, minimum width=2.2cm, minimum height=0.75cm] (answer) {Answer(s)};	   
	    
	    \draw[ultra thick, -Latex] (question.west) -- (mr) node[midway, above, xshift=-0.4cm] {Query};
	    \draw[ultra thick, -Latex, dotted] (support.west) -- (mr);
	    \draw[ultra thick, -Latex, dotted] (candidate.west) -- (mr);
	    \node[right = 0.2cm of mr] {Evidence};
	    \draw[ultra thick, -Latex, nice-blue] (mr) -- (answer.west) node[midway, below, xshift=-0.75cm] {{\color{black}Response}};
	\end{tikzpicture}
	}
	\caption{Our core abstraction, the \textsc{JTReader}. On the left, the responsibilities covered by the \textsc{Input}, \textsc{Model} and \textsc{Output} modules that compose a \textsc{JTReader} instance. On the right, the data format that is used to interact with a \textsc{JTReader} (dotted lines indicate that the component is optional).}
	\label{fig:jtr-overview}
\end{figure}
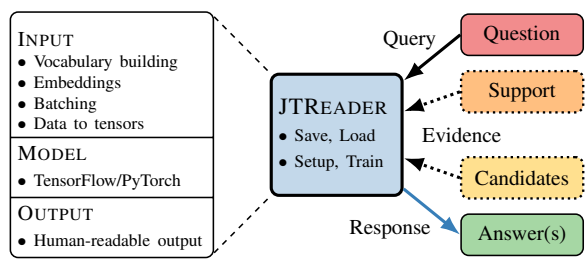
In \cref{fig:jtr-overview} we give a high-level overview of our core abstraction, the \textsc{JTReader}.
It is a task-agnostic wrapper around three typically task-dependent modules, namely the \emph{input}, \emph{model} and \emph{output} modules.
Besides serving as a container for modules, a \textsc{JTReader} provides convenience functionality for interaction, training and serialisation.
The underlying modularity is therefore well hidden from the user which facilitates the application of trained models.
\subsection{Modules and Their Usage}\label{jtr:components}
Our abstract modules have the following high-level responsibilities:
\begin{itemize}
  \item {\sc Input Modules}: Pre-processing that transforms a text-based input to tensors.
  \item {\sc Model Modules}: Implementation of the actual end-to-end MR model.
  \item {\sc Output Modules}: Converting predictions into human readable answers.
\end{itemize}
The main design for building models in {\jtr} revolves around functional interfaces between the three main modules: the input-, model-, and output module.
Each module can be viewed as a thin wrapper around a (set of) function(s) that additionally provides explicit signatures in the form of \emph{tensor ports} which can be understood as named placeholders for tensors.
The use of explicit signatures helps validate whether modules are correctly implemented and invoked, and to ensure correct behaviour as well as compatibility between modules.
Finally, by implementing modules as classes and their interaction via a simple functional interface, {\jtr} allows for the exploitation of benefits stemming from the use of object oriented programming, while retaining the flexibility offered by the functional programming paradigm when combining modules.
Given a list of training instances, corresponding to question-answer pairs, a \emph{input module} is responsible for converting such instances into tensors.
Each produced tensor is associated with a pre-defined \textit{tensor port} -- a named placeholder for a tensor -- which can in turn be used in later modules to retrieve the actual tensor.
This step typically involves some shallow forms of linguistic pre-processing such as tokenisation, building vocabularies, \etc
The \emph{model module} runs the end-to-end MR model on the now tensorised input and computes a new mapping of output tensor ports to newly computed tensors. Finally, the joint tensor mappings of the input- and model module serve as input to the \textit{output module} which produces a human-readable answer.
More in-depth documentation can be found on the project website.
\subsection{Distinguishing Features}
\paragraph{Module Reusability.}
Our shallow modularisation of readers into input-, model- and output modules has the advantage that they can be reused easily.
Most of nowadays state-of-the-art MR models require the exact same kind of input pre-processing and produce output of the same form.
Therefore, existing input- and output modules that are responsible for pre- and post-processing can be reused in most cases, which enables researchers to focus on prototyping and implementing new models.
Although we acknowledge that most of the pre-processing can easily be performed by third-party libraries such as \textsc{CoreNLP}, \textsc{NLTK} or \textsc{spaCy}, we argue that additional functionality, such as building and controlling vocabularies, padding, batching, \etc, and connecting the pre-processed output with the actual model implementation pose time intensive implementation challenges.
These can be avoided when working with one of our currently supported tasks -- Question Answering, Natural Language Inference, or Link Prediction in Knowledge Graphs.
Note that modules are typically task specific and not shared directly between tasks.
However, utilities like the pre-processing functions mentioned above and model building blocks can readily be reused even between tasks.
\paragraph{Supported ML Backends.}
By decoupling modelling from pre- and post-processing we can easily switch between backends for model implementations.
At the time of writing, {\jtr} offers support for both \textsc{TensorFlow} and \textsc{PyTorch}.
This allows practitioners to use their preferred library for implementing new MR models and allows for the integration of more back-ends in the future.
\paragraph{Declarative Model Definition.}
Implementing different kinds of MR models can be repetitive, tedious, and error-prone.
Most neural architectures are built using a finite set of basic \emph{building blocks} for encoding sequences, and realising interaction between sequences (\eg via attention mechanisms).
For such a reason, {\jtr} allows to describe these models at a high level, as a composition of simpler building blocks~\footnote{For instance, see \href{https://github.com/uclmr/jack/blob/master/conf/nli/esim.yaml}{this example}.}, leaving concrete implementation details to the framework.
The advantage of using such an approach is that is very easy to change, adapt or even create new models without knowing any implementation specifics of {\jtr} or its underlying frameworks, such as \textsc{TensorFlow} and \textsc{PyTorch}.
This solution also offers another important advantage: it allows for easy experimentation of automated architecture search and optimisation (AutoML).
{\jtr} already enables the definition of new models purely within configuration files without writing any source code.
These are interpreted by {\jtr} and support a (growing) set of pre-defined building blocks.
In fact, many models for different tasks in {\jtr} are realised by high-level architecture descriptions.
An example of an high-level architecture definition in {\jtr} is available in \cref{appx:architecture_design}.
\paragraph{Dataset Coverage.}
{\jtr} allows parsing a large number of datasets for QA, NLI, and Link Prediction.
The supported QA datasets include SQuAD~\citep{rajpurkar-EtAl:2016:EMNLP2016},
TriviaQA~\citep{DBLP:conf/acl/JoshiCWZ17},
NewsQA~\citep{DBLP:journals/corr/TrischlerWYHSBS16},
and QAngaroo~\citep{DBLP:journals/corr/abs-1710-06481}.
The supported NLI datasets include SNLI~\citep{bowman-EtAl:2015:EMNLP},
and MultiNLI~\citep{DBLP:journals/corr/WilliamsNB17}.
The supported Link Prediction datasets include WN18~\citep{DBLP:conf/nips/BordesUGWY13},
WN18RR~\citep{dettmers2018aaai}, and FB15k-237~\citep{toutanova2015observed}.
\paragraph{Pre-trained Models.}
{\jtr} offers several pre-trained models.
For QA, these include FastQA, BiDAF, and JackQA trained on SQuAD and TriviaQA.
For NLI, these include DAM and ESIM trained on SNLI and MultiNLI.
For LP, these include DistMult and ComplEx trained on WN18, WN18RR and FB15k-237.
\section{Supported MR Tasks} \label{sec:supported_tasks}
Most end-user MR tasks can be cast as an instance of question answering.
The input to a typical question answering setting consists of a \emph{question}, \emph{supporting texts} and \emph{answers} during training.
In the following we show how {\jtr} is used to model our currently supported MR tasks.
Ready to use implementations for these tasks exist which allows for rapid prototyping.
Researchers interested in developing new models can define their architecture in \textsc{TensorFlow} or \textsc{PyTorch}, and reuse existing of input- and output modules.
New datasets can be tested quickly on a set of implemented baseline models after converting them to one of our supported formats.
\paragraph{Extractive Question Answering.}
{\jtr} supports the task of \emph{Extractive Question Answering} (EQA), which requires a model to extract an answer for a question in the form of an answer span comprising a document id, token start and -end from a given set of supporting documents.
This task is a natural fit for our internal data format, and is thus very easy to represent with {\jtr}.
\paragraph{Natural Language Inference.}
Another popular MR task is \emph{Natural Language Inference}, also known as Recognising Textual Entailment (RTE).
The task is to predict whether a \emph{hypothesis} is entailed by, contradicted by, or neutral with respect to a given \emph{premise}.
In {\jtr}, NLI is viewed as an instance of multiple-choice Question Answering problem, by casting the hypothesis as the question, and the premise as the support.
The answer candidates to this question are the three possible outcomes or classes -- namely \emph{entails}, \emph{contradicts} or \emph{neutral}.

\paragraph{Link Prediction.}
A Knowledge Graph is a set of $(s, p, o)$ triples, where $s, o$ denote the \emph{subject} and \emph{object} of the triple, and $p$ denotes its \emph{predicate}: each $(s, p, o)$ triple denotes a fact, represented as a relationship of type $p$ between entities $s$ and $o$, such as: (\textsc{London}, \textsc{capitalOf}, \textsc{UK}).
Real-world Knowledge Graphs, such as Freebase~\citep{DBLP:conf/aaai/BollackerCT07}, are largely incomplete: the \emph{Link Prediction} task consists in identifying missing $(s, p, o)$ triples that are likely to encode true facts~\citep{DBLP:journals/pieee/Nickel0TG16}.
{\jtr} also supports Link Prediction, because existing LP models can be cast as multiple-choice Question Answering models, where the question is composed of three words -- a subject $s$, a predicate $p$, and an object $o$.
The answer candidates to these questions are \emph{true} and \emph{false}.
In its original formulation of the Link Prediction task, the support is left empty.
However, {\jtr} facilitates enriching the questions with additional support -- consisting, for instance, of the neighbourhood of the entities involved in the question, or sentences from a text corpus that include the entities appearing in the triple in question. Such a setup can be interpreted as an instance of NLI, and existing models not originally designed for solving Link Prediction problems can be trained effortlessly.

\section{Experiments}

Experimental setup and results for different models on the three above-mentioned MR tasks are reported in this section.
Note that our re-implementations or training configurations may not be entirely faithful.
We performed slight modifications to original setups where we found this to perform better in our experiments, as indicated in the respective task subsections.
However, our results still vary from the reported ones, which we believe is due to the extensive hyper-parameter engineering that went into the original settings, which we did not perform.
For each experiment, a ready to use training configuration as well as pre-trained models are part of {\jtr}.
\paragraph{Question Answering.}
For the Question Answering (QA) experiments we report results for our implementations of FastQA~\cite{DBLP:conf/conll/WeissenbornWS17}, BiDAF~\cite{DBLP:journals/corr/SeoKFH16} and, in addition, our own JackQA implementations.
With JackQA we aim to provide a fast and accurate QA model.
Both BiDAF and JackQA are realised using high-level architecture descriptions, that is, their architectures are purely defined within their respective configuration files.
Results of our models on the SQuAD~\cite{rajpurkar-EtAl:2016:EMNLP2016} development set along with additional run-time and parameter metrics are presented in Table~\ref{tab:QARes}.
Apart from SQuAD, {\jtr} supports the more recent NewsQA~\cite{DBLP:journals/corr/TrischlerWYHSBS16} and TriviaQA~\cite{DBLP:conf/acl/JoshiCWZ17} datasets too.

\begin{table}[t]
\small
\begin{center}
\resizebox{\columnwidth}{!}{%
\begin{tabular}{l c c c c}
\toprule
{\bf Model} & {\bf Original F1} & {\bf {\jtr} F1} & {\bf Speed} & {\bf \#Params}  \\
\midrule
BiDAF & 77.3 & 77.8 & 1.0x & 2.02M \\
FastQA & 76.3 & 77.4 & 2.2x & 0.95M \\
JackQA & -- & 79.6 & 2.0x & 1.18M \\
\bottomrule
\end{tabular}
}%
\end{center}
\caption{\label{tab:QARes} Metrics on the SQuAD development set comparing F1 metric from the original implementation to that of {\jtr}, number of parameters, and relative speed of the models.}
\vspace{-5pt}
\end{table}

\paragraph{Natural Language Inference.}
For NLI, we report results for our implementations of conditional BiLSTMs (cBiLSTM)~\cite{rocktaschel2016reasoning}, the bidirectional version of conditional LSTMs~\cite{augenstein-EtAl:2016:EMNLP2016}, the Decomposable Attention Model (DAM, \citealp{DBLP:conf/emnlp/ParikhT0U16}) and Enhanced LSTM (ESIM, \citealp{DBLP:conf/acl/ChenZLWJI17}).
ESIM was entirely implemented as a \emph{modular} NLI model, \ie its architecture was purely defined in a configuration file -- see \cref{appx:architecture_design} for more details.
Our models or training configurations contain slight modifications from the original which we found to perform better than the original setup.
Our results are slightly differ from those reported, since we did not always perform an exhaustive hyper-parameter search.
\begin{table}[t]
 \small
 \begin{center}
   \begin{tabular}{lcc}
    \toprule
    {\bf Model} & {\bf Original} & {\bf {\jtr}} \\
    \midrule
    cBiLSTM~\citep{rocktaschel2016reasoning} & -- & 82.0 \\
    DAM~\citep{DBLP:conf/emnlp/ParikhT0U16} & 86.6 & 84.6 \\
    ESIM~\citep{DBLP:conf/acl/ChenZLWJI17} & 88.0 & 87.2 \\
    \bottomrule
   \end{tabular}
 \end{center}
 \caption{\label{tab:SNLIRes} Accuracy on the SNLI test set achieved by cBiLSTM, DAM, and ESIM.}
 \vspace{-5pt}
\end{table}
\paragraph{Link Prediction.}
\begin{table}[t]
 \small
 \begin{center}
  
   \begin{tabularx}{\columnwidth}{rlccc}
    \toprule
    {\bf Dataset} & {\bf Model} & {\bf MRR} & {\bf Hits@3} & {\bf Hits@10} \\
    \midrule

    \multirow{2}{*}{\bf WN18} & DistMult & 0.822 & 0.914 & 0.936\\
    & ComplEx & 0.941 & 0.936 & 0.947\\
    
    \midrule
    
    \multirow{2}{*}{\bf WN18RR} & DistMult & 0.430 & 0.443 & 0.490 \\
    & ComplEx & 0.440 & 0.461 & 0.510 \\
    
    \midrule
    
    \multirow{2}{*}{\bf FB15k-237} & DistMult & 0.241 & 0.263 & 0.419 \\
    & ComplEx & 0.247 & 0.275 & 0.428 \\
    
    \bottomrule
  
   \end{tabularx}
 \end{center}
 \caption{\label{tab:LPRes}  Link Prediction results, measured using the Mean Reciprocal Rank (MRR) and Hits@10, for DistMult~\citep{yang15:embedding}, and ComplEx~\citep{DBLP:conf/icml/TrouillonWRGB16}.}
 \vspace{-5pt}
\end{table}
For Link Prediction in Knowledge Graphs, we report results for our implementations of DistMult~\citep{yang15:embedding} and ComplEx~\citep{DBLP:conf/icml/TrouillonWRGB16} on various datasets.
Results are outlined in \cref{tab:LPRes}.
\section{Demo}
We created three tutorial \emph{Jupyter} notebooks at \href{https://github.com/uclmr/jack/tree/master/notebooks}{this link} to demo {\jtr}'s use cases.
The quick start notebook shows how to quickly set up, load and run the existing systems for QA and NLI.
The model training notebook demonstrates training, testing, evaluating and saving QA and NLI models programmatically. However, normally the user will simply use the provided training script from command line.
The model implementation notebook delves deeper into implementing new models from scratch by writing all modules for a custom model.
\section{Conclusion}
We presented Jack the Reader ({\jtr}), a shared framework for Machine Reading tasks that will allow component reuse and easy model transfer across both datasets and domains.
{\jtr} is a new unified Machine Reading framework applicable to a range of tasks, developed with the aim of increasing researcher efficiency and code reuse.
We demonstrate the flexibility of our framework in terms of three tasks: Question Answering, Natural Language Inference, and Link Prediction in Knowledge Graphs.
With further model additions and wider user adoption, {\jtr} will support faster and reproducible Machine Reading research, enabling a building-block approach to model design and development.

\bibliographystyle{acl_natbib}
\bibliography{acl2018}

\newpage
\clearpage

\appendix

\section{High-level Architecture Design in Jack}\label{appx:architecture_design}

\newcommand\YAMLcolonstyle{\color{red}\mdseries}
\newcommand\YAMLkeystyle{\color{black}\bfseries}
\newcommand\YAMLvaluestyle{\color{blue}\mdseries}

\makeatletter

\newcommand\language@yaml{yaml}

\expandafter\expandafter\expandafter\lstdefinelanguage
\expandafter{\language@yaml}
{
  keywords={true,false,null,y,n},
  keywordstyle=\color{darkgray}\bfseries,
  basicstyle=\YAMLkeystyle,
  sensitive=false,
  comment=[l]{\#},
  morecomment=[s]{/*}{*/},
  commentstyle=\color{purple}\ttfamily,
  stringstyle=\YAMLvaluestyle\ttfamily,
  moredelim=[l][\color{orange}]{\&},
  moredelim=[l][\color{magenta}]{*},
  moredelim=**[il][\YAMLcolonstyle{:}\YAMLvaluestyle]{:},
  morestring=[b]',
  morestring=[b]",
  literate =    {---}{{\ProcessThreeDashes}}3
                {>}{{\textcolor{red}\textgreater}}1     
                {|}{{\textcolor{red}\textbar}}1 
                {\ -\ }{{\mdseries\ -\ }}3,
}

\lst@AddToHook{EveryLine}{\ifx\lst@language\language@yaml\YAMLkeystyle\fi}

We provide support for the modular composition of QA and NLI architectures within configuration files, so there is no need to touch code at all. An example configuration snippet that shows the definition of our JackQA model is presented in Listing~\ref{lst:jack_qa_light}. 
We start with a set of pre-defined start keys (`question', `char\_question', `support' and `char\_support' for QA). These refer to their respective embedded sequences. The architecture is built by a sequence of modular neural building blocks, in short \textit{modules}. Each module receives an input (a tensor or list of tensors) as determined by the given input keys and produces an output which can be referred to in subsequent modules using the provided output key. In case no output key is given, it defaults to the given input key or the first of a list of given input keys.
More detailed information can be found in our online documentation.

\begin{lstlisting}[language=yaml, basicstyle=\small, caption={Sample \textsc{YAML} architecture description for our JackQA model.}, captionpos=b, frame=single, label={lst:jack_qa_light}]
repr_dim: 100

model:
  encoder_layer:
  # Shared encoding
  # Support
  - input: ['support', 'char_support']
    output: 'support'
    module: 'concat'
  - input: 'support'
    name: 'embedding_highway'
    module: 'highway'
  - input: 'support'
    output: 'emb_support'
    name: 'embedding_projection'
    module: 'dense'
    activation: 'tanh'
    dropout: True
  - input: 'emb_support'
    output: 'support'
    module: 'conv_glu'
    conv_width: 5
    num_layers: 2
    name: 'contextual_encoding'
    dropout: True
  - input: ['emb_support', 'support']
    output: 'enc_support'
    module: 'concat'
    
  # Question
  - input: ['question', 'char_question']
    output: 'question'
    module: 'concat'
  - input: 'question'
    name: 'embedding_highway'
    module: 'highway'
  - input: 'question'
    output: 'emb_question'
    name: 'embedding_projection'
    module: 'dense'
    activation: 'tanh'
    dropout: True
  - input: 'emb_question'
    output: 'question'
    module: 'conv_glu'
    conv_width: 5
    num_layers: 2
    name: 'contextual_encoding'
    dropout: True
  - input: ['emb_question', 'question']
    output: 'enc_question'
    module: 'concat'

  # Attention
  - input: 'enc_support'
    dependent: 'enc_question'
    output: 'support'
    module: 'attention_matching'
    attn_type: 'diagonal_bilinear'
    with_sentinel: True
    scaled: True
  - input: 'support'
    output: 'support_self'
    module: 'dense'
    activation: 'tanh'

  # Self Attention
  - input: 'support_self'
    module: 'self_attn'
    attn_type: 'diagonal_bilinear'
    scaled: True
    with_sentinel: True
    num_attn_heads: 1

  # Concatenate outputs
  - input: ['support', 'support_self']
    output: 'support'
    module: 'concat'
  - input: 'support'
    module: 'dense'
    activation: 'relu'
    dropout: True

  # BiLSTM, the only application
  - input: 'support'
    module: 'lstm'
    with_projection: True
    activation: 'tanh'
    dropout: True

  - input: 'support'
    module: 'conv_glu'
    conv_width: 5
    num_layers: 1
    residual: True

  answer_layer:
    support: 'support'
    question: 'enc_question'
    module: 'bilinear'
\end{lstlisting}

\end{document}


\maketitle

\appendix

\section{Using {\jtr} as a Framework}
~\\

\begin{figure}[h]

\begin{lstlisting}[language=jtr,firstnumber=1,style=nonumbers,basicstyle=\small\ttfamily,columns=fullflexible]
# Create a reader
from jtr.preprocess.vocab import Vocab
embedding_dim = hidden_dim = 128
config = {'batch_size': 128,
          'repr_dim': hidden_dim,
          'repr_dim_input': embedding_dim}
reader = readers.readers['snli_reader'](
        Vocab(), config)

# Hooks to monitor loss and metrics
from jtr.jack.train.hooks import 
        (LossHook,
        ClassificationEvalHook)
hooks = [LossHook(reader,iter_interval=10),
         readers.eval_hooks['snli_reader']
         (reader, dev_set, iter_interval=25)]
         
# Initialise optimiser
import tensorflow as tf
optim = tf.train.AdamOptimizer()

# Train reader
reader.train(optim, train_set,
             hooks=hooks, max_epochs=2)

# Analyse errors
from jtr.jack.tasks.mcqa.simple_mcqa \
        import MisclassificationOutputModule
# Error analysis where the model
# predicted a probability in [0.0, 0.2]
# We want to print 10 examples
reader.output_module = \
        MisclassificationOutputModule(
            interval=[0.0, 0.20], limit=10)
reader.process_outputs(test_set)
\end{lstlisting}
\caption{Using {\jtr} to train and evaluate an existing model}
\label{lst:jtr}
\end{figure}

\vfill\null
\newpage

\section{Data examples}\label{lab:dataex}
Here, we show one data instance each for {\squad}, SNLI and NYT converted to \jtr format, which is defined as a JSON-based format.
Note that for NYT, we only show part of the candidates, as the candidate list is the same for all questions and thus very long.

\paragraph{{\squad}}
~\\

{
\begin{lstlisting}[language=json,firstnumber=1,style=nonumbers,label={lst:squad}]
{"instances": [
  {"questions": [
    "question": 
      "Where did Super Bowl 50 take place?"
    {"answers": [
      {"text": " Santa Clara, 
       California",
       "span": [402, 426]}]
       },
      {"text": "Levi's Stadium",
       "span": [355, 369]}]
       }, 
       {"text": "Levi's Stadium",
        "span": [355, 369]}]
       },
       {"text":" "Levi's Stadium in the San Francisco Bay Area at Santa Clara, California.",
        "span": [354, 427]}]
       }, 
    }]
  "support": ["Super Bowl 50 was an American football game to determine the champion of the National Football League (NFL) for the 2015 season. The American Football Conference (AFC) champion Denver Broncos defeated the National Football Conference (NFC) champion Carolina Panthers 24-10 to earn their third Super Bowl title. The game was played on February 7, 2016, at Levi's Stadium in the San Francisco Bay Area at Santa Clara, California. As this was the 50th Super Bowl, the league emphasized the \"golden anniversary\" with various gold-themed initiatives, as well as temporarily suspending the tradition of naming each Super Bowl game with Roman numerals (under which the game would have been known as \"Super Bowl L\"), so that the logo could prominently feature the Arabic numerals 50."]}],
  "meta": "Squad",
}
\end{lstlisting}
}

\vfill\null
\newpage

\paragraph{SNLI}
~\\

{
\begin{lstlisting}[language=json,firstnumber=1,style=nonumbers,label={lst:snli}]
{"instances": [
  {"questions": [
    "question": 
      "There are children present"
    {"answers": [
      {"text": 
        "entailment"}],
    }]
  "support": ["Children smiling and waving at camera"]}],
  "meta": "SNLI",
}
\end{lstlisting}
}

\paragraph{NYT}
~\\

{
\begin{lstlisting}[language=json,firstnumber=1,style=nonumbers,label={lst:nyt}]
{"instances": [
  {"questions": [
    "question": 
      "REL\$/film/film/directed_by"
    {"answers": [
      {"text": 
        "(Girlhood|||Liz Garbus)"}],
    },
    "question": 
      "path#poss|<-poss<-film->dep->|dep:INV"
    {"answers": [
      {"text": 
        "(Girlhood|||Liz Garbus)"}],
    }]
  "support": []}],
  "meta": "NAACL2013",
  "globals": {
    "candidates": [
      {"text": "(Barnaby|||Rangers)"},
      {"text": "(Australia|||Japan)"},
      {"text": "(Girlhood|||Liz Garbus)"},
      {"text": "(Zimmer|||Pennsylvania)"}
    }
}
\end{lstlisting}
}